\def\year{2021}\relax
\documentclass[letterpaper]{article} 
\usepackage{aaai21}  
\usepackage{times}  
\usepackage{helvet} 
\usepackage{courier}  
\usepackage[hyphens]{url}  
\usepackage{graphicx} 
\usepackage{graphicx}  
\usepackage{natbib}  
\usepackage{caption} 
\usepackage{booktabs}
\usepackage{subcaption}
\usepackage{algorithm}
\usepackage{algorithmic}
\usepackage{multirow}

\usepackage{color}
\definecolor{ao}{rgb}{0.0, 0.5, 0.0}
\definecolor{armygreen}{rgb}{0.29, 0.33, 0.13}
\definecolor{nkc}{rgb}{0.36, 0.54, 1}
\usepackage{xspace}
\newcommand{\mname}{\texttt{STELAR}\xspace }

\usepackage{booktabs}

\def\bbe{\boldsymbol{\beta}}

\def\bga{\boldsymbol{\gamma}}

\def\bPh{\mathbf{\Phi}}

\def\bPh{\mathbf{\Phi}}

\def\A{\mathbf{A}}

\def\B{\mathbf{B}}

\def\C{\mathbf{C}}

\def\P{\mathbf{P}}

\def\Q{\mathbf{Q}}

\def\bR{\mathbb{R}}

\def\X{\mathbf{X}}

\def\diag{\mathrm{diag}}

\def\tX{\underline{\X}} 

\usepackage{amsmath}
\usepackage{amssymb}
\DeclareMathOperator*{\argmin}{arg\,min}

\title{\mname: Spatio-temporal Tensor Factorization with Latent Epidemiological Regularization}
\author{
	Nikos Kargas\textsuperscript{\rm 1,2,\thanks{equal contribution}}, 
	Cheng Qian\textsuperscript{\rm 2,*}, 
	Nicholas D. Sidiropoulos\textsuperscript{\rm 3}, 
	Cao Xiao\textsuperscript{\rm 2},  \\ 
	Lucas M. Glass\textsuperscript{\rm 2,4}, 
	Jimeng Sun\textsuperscript{\rm 5} \\
}
\affiliations{
     \textsuperscript{\rm 1} Dept. of ECE, University of Minnesota, \textsuperscript{\rm 2} Analytics Center of Excellence, IQVIA, \textsuperscript{\rm 3} Dept. of ECE, University of Virginia, \\ 
     \textsuperscript{\rm 4} Dept. of Statistics, Temple University \textsuperscript{\rm 5} Dept. of CS, University of Illinois Urbana-Champaign 
}

\begin{document}


\maketitle
\begin{abstract}
Accurate prediction of the transmission of epidemic diseases such as COVID-19 is crucial for implementing effective mitigation measures. In this work, we develop a tensor method to predict the evolution of epidemic trends for many regions simultaneously. We construct a $3$-way spatio-temporal tensor (location, attribute, time) of case counts and propose a nonnegative tensor factorization with latent epidemiological model regularization named \mname. Unlike standard tensor factorization methods which cannot predict slabs ahead, \mname enables long-term prediction by incorporating latent temporal regularization through a system of discrete-time difference equations of a widely adopted epidemiological model. We use {\em latent} instead of location/attribute-level epidemiological dynamics to capture common epidemic profile sub-types and improve collaborative learning and prediction. We conduct experiments using  both county- and state-level COVID-19 data and show that our model can identify interesting latent patterns of the epidemic. Finally, we evaluate the predictive ability of our method and show superior performance compared to the baselines, achieving up to $21\%$ lower root mean square error and $25\%$ lower mean absolute error for county-level prediction.
\end{abstract}

\section{Introduction}
\label{intro}
Pandemic diseases such as the novel coronavirus disease (COVID-19) pose a serious threat to global public health, economy, and daily life. Accurate epidemiological measurement, modeling, and tracking are needed to inform public health officials, government executives, policy makers, emergency responders, and the public at large.  
Two types of epidemiological modeling methods are popular today:
\begin{itemize}
    \item {\bf Mechanistic models} define a set of ordinary differential equations (ODEs) which capture the epidemic transmission patterns and predict the long-term trajectory of the outbreak. These models include the Susceptible-Infected-Recovered (SIR) model~\cite{kermack1927}, the Susceptible-Exposed-Infected-Recovered (SEIR)~\cite{cooke1996} and their variants e.g, SIS, SIRS, and delayed SIR. These models have a small number of parameters which are determined via curve fitting. These types of models do not require much training data, but they are quite restrictive and cannot leverage rich information. 
    \item {\bf Machine learning models} such as deep learning, model the epidemiological trends as a set of regression problems~\cite{yang2020,toda2020,he2020,chimmula2020,tomar2020}. These models are able to learn the trajectory of the outbreak only from data and can perform very well especially in short-term prediction. However, they usually require a large amount of training data. 
\end{itemize}

This paper extends Canonical Polyadic Decomposition (CPD)~\cite{Harshman1970} via epidemiological model (e.g., SIR and SEIR) regularization, which integrates spatio-temporal real-world case data and exploits the correlation between different regions for long-term epidemic prediction. CPD is a powerful tensor model with successful  applications in many fields~\cite{papalexakis2016, Sidiropoulos2017}. Compared to traditional epidemiological models which cannot incorporate fine-grain observations or any kind of side information, tensors offer a natural way of representing multidimensional time evolving data and incorporate additional information~\cite{acar2009, de2017}. CPD can capture the inherent correlations between the different modes and thanks to its uniqueness properties it can extract interpretable latent components rendering it an appealing solution for modeling and analyzing epidemic dynamics. Additionally, it can parsimoniously represent multidimensional data and can therefore learn from limited data. These two important properties differentiate CPD from neural networks and deep learning models which typically require a lot of training data and are often treated as black box models. 

In this work, we propose a Spatio-temporal Tensor factorization with EpidemioLogicAl Regularization (\mname). \mname combines a nonnegative CPD model with a system of discrete-time difference equations to capture the epidemic transmission patterns. Unlike standard tensor factorization methods which cannot predict slabs ahead, \mname can simultaneously forecast the evolution of the epidemic for a list of regions, and it can also perform long-term prediction.

In the experiments, we build a spatio-temporal tensor, $\mathrm{location}\times \mathrm{attribute}\times \mathrm{time}$ of case counts based on large real-world medical claims datasets, where the first dimension corresponds to different counties/states, the second dimension corresponds to different attributes or signals that evolve over time, such as daily new infections, deaths, number of hospitalized patients and other COVID-19 related signals, and the third is the time-window of the available signals. This spatio-temporal tensor is factorized using the proposed low-rank nonnegative tensor factorization with epidemiological model regularization \mname. We observed that the extracted latent time components provide intuitive interpretation of different epidemic transmission patterns which traditional epidemiological models such as SIR and SEIR are lacking.

Our main contributions are summarized as follows:
\begin{itemize}
\item We propose \mname, a new data-efficient tensor factorization method regularized by a disease transmission model in the latent domain to predict future slabs. We show that by jointly fitting a low-rank nonnegative CPD model with an SIR model on the time factor matrix we are able to accurately predict the evolution of epidemic trends.
\item Thanks to the uniqueness properties of the CPD, our method produces interpretable prediction results. Specifically, the tensor is approximated via $K$ rank-1 components, each of which is associated with a sub-type of the epidemic transmission patterns in a given list of regions. The latent time factor matrix includes $K$ different patterns of the epidemic evolution and using the latent location and signal factor matrices we can identify the corresponding locations and signals associated with each pattern.
\item We perform extensive experimental evaluation on both county- and state-level COVID-19 case data for $10$ and $15$ days-ahead prediction and show that our method outperforms standard epidemiological and machine learning models in COVID-19 pandemic prediction. Our method achieves up to $21\%$ lower root mean square error and $25\%$ lower mean absolute error for county-level prediction.
\end{itemize}

\begin{figure*}[t]
\centering
\begin{subfigure}{0.22 \textwidth}
\includegraphics[width=1\textwidth]{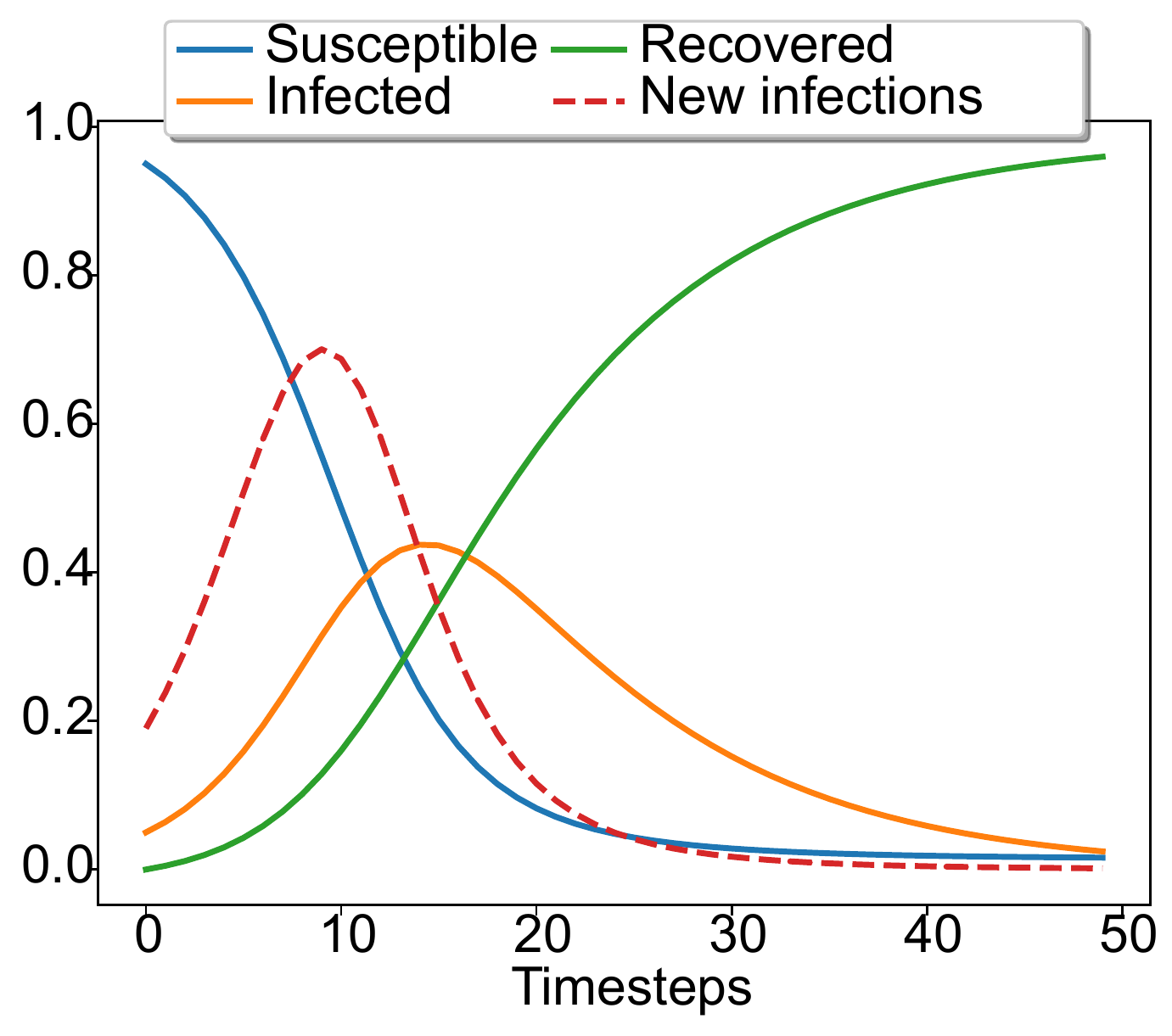}
\caption{SIR model.}
\label{fig:sir}
\end{subfigure}
\begin{subfigure}{0.22 \textwidth}
\includegraphics[width= 1\textwidth]{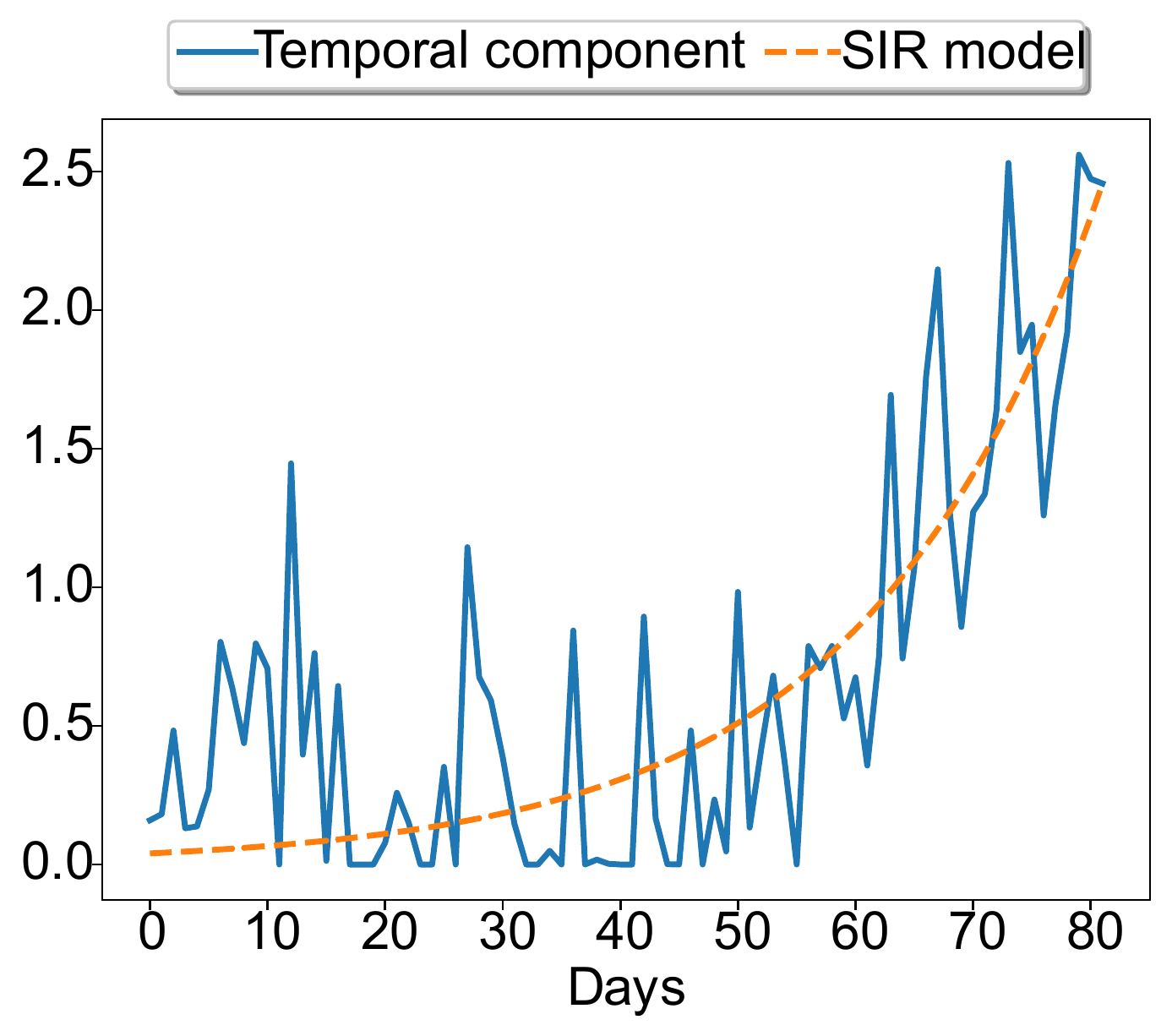}
\caption{Component $1$.}
\label{fig:component1}
\end{subfigure}
\begin{subfigure}{0.22 \textwidth}
\includegraphics[width=1\textwidth]{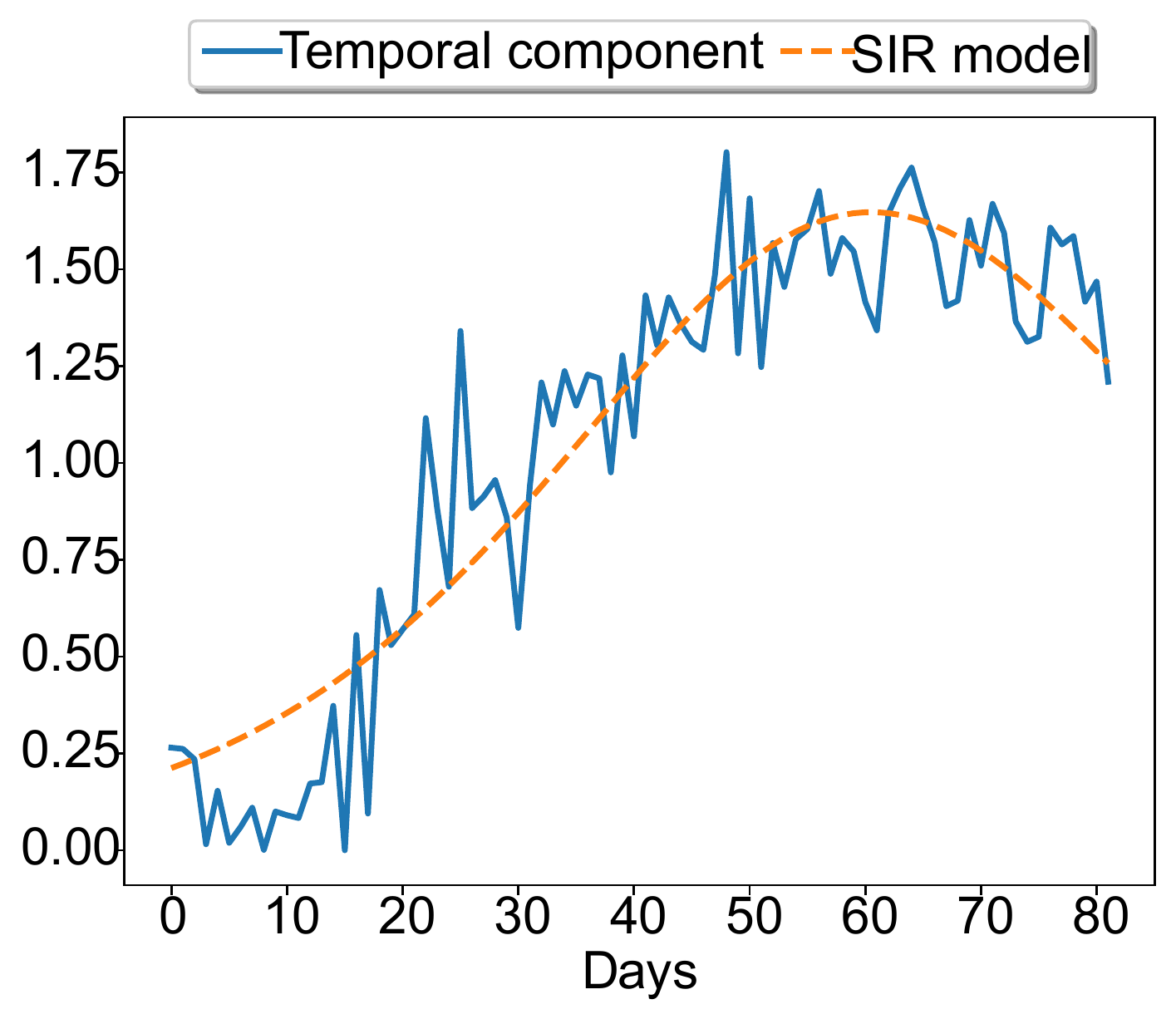}
\caption{Component $2$.}
\label{fig:component2}
\end{subfigure}
\begin{subfigure}{0.22\textwidth}
\includegraphics[width=1\textwidth]{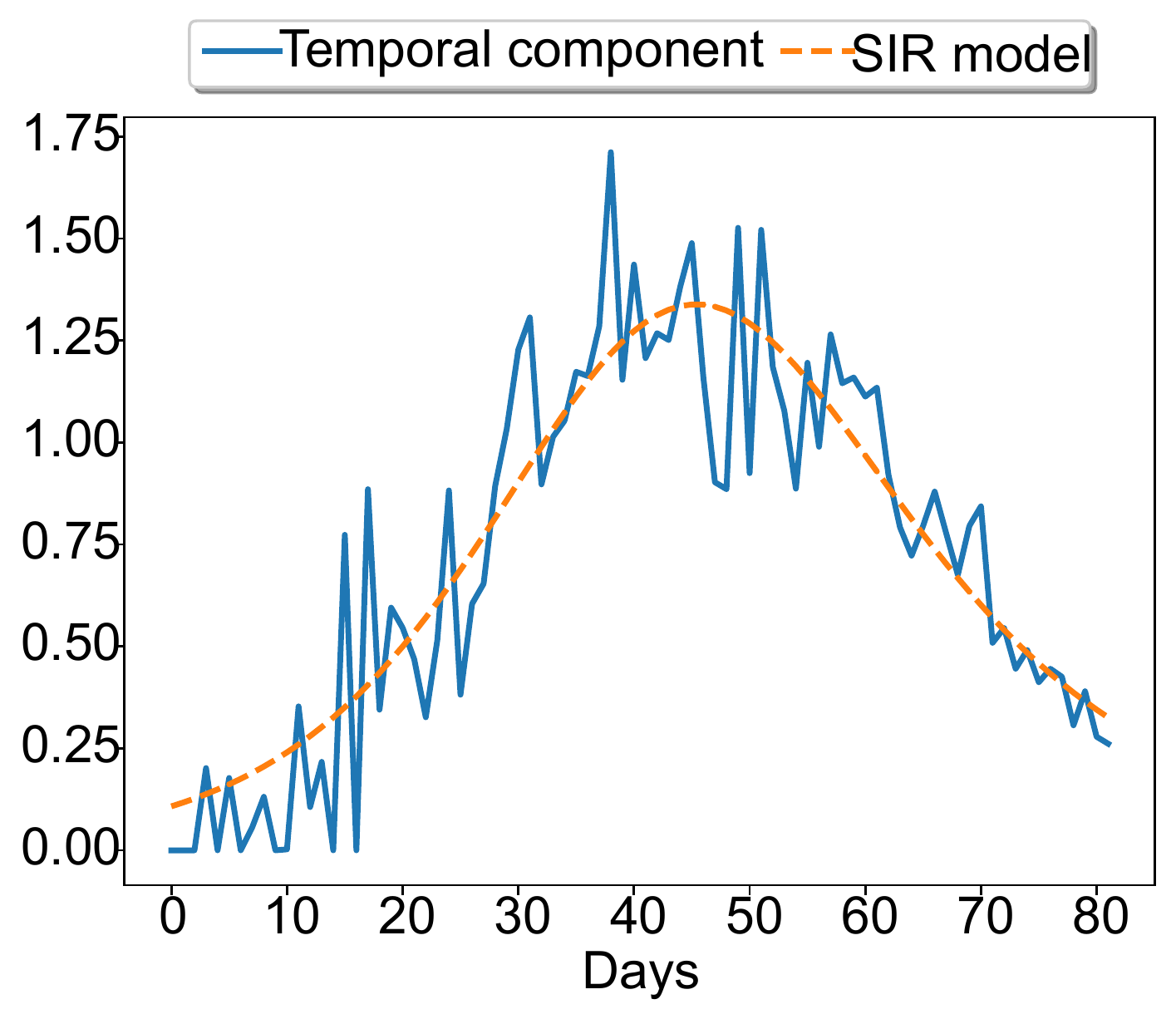}
\caption{Component $3$.}
\label{fig:component3}
\end{subfigure}
\caption{An example of an SIR model with parameters $N=1, S(0) = 0.95, I(0) = 0.05, R(0) =0, \beta =0.4, \gamma = 0.1 $ is shown in Figure~\ref{fig:sir}. The new infections curve has been scaled by $10$ for visualization purposes. Figures~\ref{fig:component1},~\ref{fig:component2},~\ref{fig:component3} show $3$ latent time components i.e., columns of the factor matrix $\C$ when we compute a plain nonnegative CPD of the spatio-temporal tensor.}
\end{figure*}

\section{Related Work}

Many epidemiological and deep learning models have been applied for modeling the COVID-19 pandemic evolution. Methods based on traditional epidemic prediction models such as SIR~\cite{kermack1927} and SEIR and their variants~\cite{cooke1996}, rely on a system of differential equations which describe the dynamics of the pandemic~\cite{yang2020,toda2020,he2020}. These models are trained for each location independently by curve fitting.

Machine learning approaches and specifically deep learning models treat the problem as a time series prediction problem. Given as input a  time series of length $L_w$, $x_{t-L_w-1},\ldots,x_{t-1},x_{t}$, the goal is to output $L_o$ future time points $x_{t+1},x_{t+2},\ldots,x_{L_o}$. A popular and very successful method for time series prediction is the Long Short Term Memory (LSTM) network~\cite{hochreiter1997}. Recent studies have applied LSTMs for COVID-19 prediction~\cite{chimmula2020, tomar2020, yang2020}. To increase the data for LSTM training,~\citeauthor{yang2020} used the 2003 SARS-CoV data to pretrain the model before using it for COVID-19 prediction. DEFSI~\cite{wang2019} is a deep learning method based on LSTMs for influenza like illness forecasting which combines synthetic epidemic data with real data to improve epidemic forecasts. 

Graph Neural Network (GNN)~\cite{kipf2017} is a type of neural network which operates on a graph.~\citeauthor{kapoor2020} proposed creating a graph with spatial and temporal edges. They leverage  human mobility data to improve the prediction of daily new infections~\cite{kapoor2020}. STAN~\cite{gao2020} is an attention-based graph convolutional network which constructs edges  based on geographical proximity of the different regions and regularizes the model predictions based on an epidemiological model.

Methods based on tensor factorization have been applied for various time series prediction tasks. CP Forecasting~\cite{dunlavy2011}, is a tensor method which computes a low-rank CPD model and uses the temporal factor to capture periodic patterns in data. TENSORCAST~\cite{de2017} is a method that forecasts time-evolving networks using coupled tensors. Both methods are not suitable for COVID-19 pandemic prediction and rely on a two-step procedure -- fitting a low-rank model and then performing forecasting based on the temporal factor matrix -- which as we will see later leads to performance degradation relative to our joint optimization formulation. 
\begin{table}[t]
    \centering
    \begin{tabular}{ c|l}
         \toprule
         Notation & Description\\
         \midrule
         $\tX\in\bR^{M\times N \times L}$  & spatio-temporal tensor \\
         $\A\in\bR^{M\times K}$   & location factor matrix \\
         $\B\in\bR^{N\times K}$ &  signal factor matrix\\
         $\C\in\bR^{L\times K}$ & temporal factor matrix\\
         $\tX^{(i)}$ & mode-$i$ unfolding of $\tX$  \\
         $\beta$; $\gamma$; & contact rate; recovery rate\\
         $K$ & \# of components \\
         $M$ & \# of locations \\
         $N$ & \# of signals \\
         $L$ & \# of time points\\
         $T$ & transpose \\
         $\circ$ & outer product\\
         $\otimes$ & Kronecker product \\
         $\odot$ & Khatri-Rao product\\
         $\circledast$ & Hadamard product\\
         $\|\cdot\|_F$ & Frobenius norm \\
         $\rm{diag}(\mathbf{x})$ & diagonal matrix\\ 
         \bottomrule
    \end{tabular}
    \caption{Notation}
    \label{table:notation}
\end{table}

\section{Background}
In this section, we review necessary background on epidemiological models and tensor decomposition that will prove useful for developing our method. Table~\ref{table:notation} contains the notation used throughout the paper.

\subsection{Epidemiological Models} 
Epidemiological models have been popular solutions for pandemic modeling. For example, The SIR model~\cite{kermack1927} is one of the most famous and paradigmatic models in mathematical epidemiology. In this model, a population is divided into susceptible, infected and recovered subpopulations. The evolution of these quantities is described by the following equations:
\begin{subequations}
\begin{align}
& S(t) - S(t-1) = - \beta S(t-1) I(t-1)/N,  \label{eq:dS}                     \\
& I(t) - I(t-1) =  \frac{\beta S(t-1) I(t-1)}{N}- \gamma I(t-1), \label{eq:dI}\\
& R(t) - R(t-1) = \gamma I(t-1), \label{eq:dR}
\end{align}
\label{eq:sir}
\end{subequations}
where $S(t)$, $I(t)$ and $R(t)$ stand for the size of susceptible, infected and recovered populations at time $t$, respectively and  $N = S(t) + I(t) + R(t) $  is the total population. The parameter $\beta$ controls the rate of spread and the parameter $\gamma$ is the recovery rate, so its inverse $1/\gamma$ represents the average time period that an infected individual remains infectious. In Equation~\eqref{eq:dS}, the quantity $\beta S(t) / N$ is the fraction of susceptible individuals that will contact an infected individual at time $t$. Therefore,
\begin{equation}
C(t) := \beta S(t)I(t)/N,
\label{eq:new_infections}
\end{equation}
denotes the new infections at time $t$. In Equation~\eqref{eq:dI}, $\gamma I(t)$ denotes the number of individuals recovered. Therefore, the change of the infected cases at time $t$ will be the difference between new infected and recovered cases as shown in Equation~\eqref{eq:dI}. Given $S(0)$, $I(0)$ and $R(0)$, we can compute $S(t)$, $I(t)$ and $R(t)$ for $t=1,2,\ldots$, $L$, where $L$ is the size of the prediction window we are interested in. An example is shown in Figure~\eqref{fig:sir} where we have set $N=1, S(0) = 0.95, I(0) = 0.05, R(0) =0, \beta =0.4, \gamma = 0.1 $ i.e., we have normalized the different populations by the total population number $N$ and set the window size to be $L=50$. 

The SEIR model is a variant of the SIR model. In this model, a population is divided into susceptible, exposed, infected and recovered subpopulations. The exposed population consists of individuals who have been infected but are not yet infectious. The SEIR model includes an additional parameter $\sigma$ which is the rate at which the exposed individuals becoming infectious.

\subsection{Canonical Polyadic Decomposition}
The Canonical Polyadic Decomposition (CPD) expresses a \mbox{$3$-way} tensor $\tX \in \mathbb{R}^{M \times N \times L}$ as a sum of \mbox{rank-$1$} components, i.e.,
\begin{equation}
\tX = [\![\A, \B, \C]\!] = \sum_{k=1}^K\mathbf{a}_k \circ \mathbf{b}_k \circ \mathbf{c}_k,
\label{eq:cpd}
\end{equation}
where $\mathbf{A} = [ \mathbf{a}_1, \ldots, \mathbf{a}_K] \in \mathbb{R}^{M \times K}$, $\mathbf{B} = [ \mathbf{b}_1, \ldots, \mathbf{b}_K] \in \mathbb{R}^{N \times K}$, $\mathbf{C} = [ \mathbf{c}_1, \ldots, \mathbf{c}_K] \in \mathbb{R}^{L \times K}$. The rank $K$ is the minimum number of components needed to synthesize $\tX$. We can express the CPD of a tensor in many different ways. For example, $\tX = [\![\A, \B, \C]\!]$ can be represented by the matrix unfolding $\tX^{(1)} = (\C \odot \B) \A^T $ where the mode-$1$ fibers are the rows of the resulting matrix. Using ‘role symmetry’, the mode-$2$ and mode-$3$ matrix unfoldings are given by  $\tX^{(2)} = (\C \odot \A) \B^T $,  $\tX^{(3)} = (\B \odot \A) \C^T $  respectively. CPD can parsimoniously represent tensors of size $M \times N \times L$ using only $(M+N+L)\times K$ parameters. The CPD model has two very important properties that make it a very powerful tool for data analysis 1) it is universal, i.e., every tensor admits a CPD of finite rank, and 2) it is unique under mild conditions i.e., it is possible to extract the true latent factors that synthesize $\tX$. 

\section{Method}\label{method}

\subsection{Problem Formulation}
Consider a location where we monitor $N$ signals related to a pandemic over time, e.g., number of new infections, hospitalized patients, intensive care unit (ICU) patients, etc. At time $t$, the value of the $n$th signal at location $m$ is denoted by $x_{m,n,t}$. Assuming that there are $N$ signals, $M$ locations and $L$ time points, then, the dataset can be naturally described by a $3$-way spatio-temporal tensor $\tX \in \bR^{M \times N \times L}$, where $\tX(m,n,t) := x_{m,n,t}$. Tensor $\tX$ includes the evolution of all signals and regions for times $1$ through $L$ and we are interested in estimating the signals at time $L+1,\ldots,L+L_o$. In other words, we want to predict the frontal slabs $\tX(:,:,t)$ for $L_o$ timesteps ahead. However, standard  tensor  factorization methods cannot predict slabs ahead. It is evident from Equation~\ref{eq:cpd} and the mode-$3$ unfolding that is impossible to impute when an entire slab $\tX(:,:,t)$ is missing since we have no information regarding the corresponding row of $\C$. To address this challenge, we take into consideration the transmission law dynamics of the disease. The key idea is to decompose the tensor using a CPD model and impose SIR constraints on the latent time factor.

To illustrate the key idea, we perform a preliminary experiment using a  rank-$5$ plain nonnegative CPD on a spatio-temporal tensor $\tX \in \mathbb{R}^{140 \times 15 \times 82}$ with case counts, constructed from real COVID-19 data. As shown in Figures~\ref{fig:component1},~\ref{fig:component2},~\ref{fig:component3}, CPD is able  to  extract meaningful latent components from the data. Specifically, the figures show $3$ columns of the latent time factor $\C$ where an SIR model has been fitted after obtaining the decomposition. We observe that each columns depicts a curve similar to the one in Figure~\ref{fig:sir} i.e., the CPD can unveil the principal patterns of the epidemic evolution, and each pattern corresponds to a different pandemic phase. 

Therefore, we propose solving the following constrained nonnegative CPD problem
\begin{align}\label{eq:model}
\min_{ \substack{ \A ,\B,\C, \\ \bbe, \bga, \mathbf{s}, \mathbf{i}}} ~& \left \| \tX - [\![ \A, \B, \C]\!] \right \|_F^2 + \mu  \left( \| \A\|_F^2 + \| \B\|_F^2+ \| \C\|_F^2  \right)   \notag\\
 & + \nu \sum_{k=1}^K \sum_{t=1}^L \left (c_{t,k} - \beta_k S_k(t-1) I_k(t-1) \right)^2 \notag\\
 \text{s. t.} ~~&\A \geq \mathbf{0}, \B \geq \mathbf{0},\C \geq \mathbf{0},  \notag \\
 & \bbe \geq \mathbf{0}, \bga \geq \mathbf{0}, \mathbf{s} \geq \mathbf{0},  \mathbf{i} \geq \mathbf{0}, \\
& S_k(t) = S_k(t-1) - \beta_k S_k(t-1) I_k(t-1), \notag\\
& I_k(t) = I_k(t-1) + \beta_k S_k(t-1) I_k(t-1) \notag\\
&\qquad \quad - \gamma_k I_k(t-1), \notag \\
& s_k = S_k(0), i_k = I_k(0) \notag. 
\end{align}
The first term  is the data fitting term. We fit a CPD model of rank-$K$ with nonnegativity constraints on the factors. The second term is Frobenius norm regularization which is typically used to avoid overfitting and improve generalization of the model. We introduce a third term which regularizes each column of the factor matrix $\C$ according to Equation~\eqref{eq:new_infections} i.e., we learn $K$ different SIR models, each of which is fully described by parameters $S_k(0), I_k(0), R_k(0)$ and $\beta_k$, $\gamma_k$ according to Equations~\eqref{eq:dS},~\eqref{eq:dI},~\eqref{eq:dR}. We aim at estimating these parameters such that each column of $\C$  follows the new infections curve of an SIR model. Vectors $\bbe \in \mathbb{R}^K$ and $\bga \in \mathbb{R}^K$ hold the $\beta$ and $\gamma$ parameters of each SIR model and vectors $\mathbf{s} = \in \mathbb{R}^K $, $\mathbf{i} \in \mathbb{R}^K$ hold the initial conditions of each SIR model. Note that the recovered cases curve can be safely removed because it does not affect the optimization cost and that the parameter $N_k$ corresponding to the total population has been absorbed in the vector $\boldsymbol{\beta}$.

\subsection{Prediction}
After the convergence of the above optimization algorithm, we have some estimates of $\hat\A$, $\hat\B$, $\hat\C$ and parameters $\{\beta_1, \cdots, \beta_K\}$, $\{\gamma_1, \cdots, \gamma_K\}$, $\{s_1, \cdots, s_K\}$, $\{i_1, \cdots, i_K\}$ where the pair of $\beta_k$ and $\gamma_k$ describes the epidemic transmission of the $k$th component in the time factor matrix and $s_k$, $i_k$ the initial values of the subpopulations. Using Equations~\eqref{eq:dS},\eqref{eq:dI} and \eqref{eq:new_infections} we can predict ``future'' values for the $k$th column of $\C$. We repeat the same procedure for all columns of $\C$ such that we can predict the entire ``future'' rows. Let $\hat{\C}(t, :) \in \mathbb{R}^{K}$ be the prediction of the temporal information at a future time point $t$ using estimates $\hat\bbe$, $\hat\bga$, $\hat{\boldsymbol{s}}, \hat{\boldsymbol{i}}$. Since $\A$ and $\B$ do not depend on $t$, the prediction of all signals in the tensor at time $t$ is given by
\begin{align}
   \widehat{\tX}(:,:,t) = \hat\A\diag(\hat{\C}(t, :) \hat\B^T.
   \label{eq:prediction}
\end{align}

Adding latent temporal regularization through an SIR model offers significant advantages compared to having separate SIR models for each location. Our model can capture correlations between different locations and signals through their latent representations and therefore improve the prediction accuracy. Additionally, it enables expressing the evolution of a signal as weighted sum of
$K$ separate SIR models e.g., the prediction of the $n$th signal for the $m$th location for time point $t$, is given by
\begin{equation}
\widehat{\tX}(m,n,t) = \sum_{k=1}^K \hat{a}_{m,k} \hat{b}_{n,k} \hat{\beta}_k S_k(t-1) I_k(t-1),
\end{equation}
which makes it much more flexible and expressive.  

\subsection{Optimization}
The optimization problem~\eqref{eq:model} is a nonconvex and very challenging optimization problem. To update factor matrices $\A,\B,\C$ we rely on alternating optimization. Note that by fixing all variables except for $\A$, the resulting subproblem w.r.t. $\A$, is a nonnegative least squares problem, which is convex. Similarly for the factor matrices $\B$ and $\C$. We choose to solve each factor matrix subproblem via the Alternating Direction Method of Multipliers (ADMM)~\cite{gabay1976}, which is a very efficient algorithm that has been successfully applied to nonnegative tensor factorization problems~\cite{huang2016}.

Let us first consider the subproblem w.r.t. $\A$. Assume that at the $\ell$th iteration, we have some estimates of all the variables available. Fixing all variables except for $\A$, we have
\begin{equation}
\min_{ \A \geq \mathbf{0}}  \| \tX^{(1)} - \bPh_A^{(\ell)}\A^T  \|_F^2  + \mu   \| \A \|_F^2
\label{eq:subA}
\end{equation}
where $\bPh_A^{(\ell)} = \C^{(\ell)} \odot \B^{(\ell)}$. The ADMM updates for optimization problem~\eqref{eq:subA} are the following
\begin{subequations}
\begin{align}
& \widehat{\mathbf{A}}  = \argmin   \| \tX^{(1)} - \bPh_A^{(\ell)} \widehat{\mathbf{A}}  \|_F^2 + \mu   \| \widehat{ \mathbf{A}} \|_F^2  \label{eq:admm1}\\ &   \quad \quad \quad + \rho  \| \mathbf{A} - \widehat{\mathbf{A}}^T +  \mathbf{A}_d  \|_F^2, \notag\\
& \mathbf{A} =  \argmin_{\mathbf{A} \geq \mathbf{0} } \rho  \| \mathbf{A} - \widehat{\mathbf{A}}^T +  \mathbf{A}_d  \|_F^2, \label{eq:admm2} \\
& \mathbf{A}_d  =  \mathbf{A}_d + \mathbf{A} - \widehat{\mathbf{A}}^T.  \label{eq:admm3} 
\end{align}
\label{eq:admmA}
\end{subequations}
Equation~\eqref{eq:admm1} is a least squares problem. Because ADMM is an iterative algorithm and the update is performed multiple times, we save computations by caching~\cite{huang2016}. Equation~\eqref{eq:admm2} is a simple element-wise nonnegative projection operator and Equation~\eqref{eq:admm3} is the dual variable update. The updates for $\B$ are similar.
\begin{subequations}
\begin{align}
& \widehat{\mathbf{B}}  = \argmin   \| \tX^{(2)} - \bPh_B^{(\ell)} \widehat{\mathbf{B}}  \|_F^2 + \mu   \| \widehat{ \mathbf{B}} \|_F^2 \\ &   \quad \quad \quad + \rho  \| \mathbf{B} - \widehat{\mathbf{B}}^T +  \mathbf{B}_d  \|_F^2, \notag\\
& \mathbf{B} =  \argmin_{\mathbf{B} \geq \mathbf{0} } \rho  \| \mathbf{B} - \widehat{\mathbf{B}}^T +  \mathbf{B}_d  \|_F^2, \\
& \mathbf{B}_d  =  \mathbf{B}_d + \mathbf{B} - \widehat{\mathbf{B}}^T.
\end{align}
\label{eq:admmB}
\end{subequations}
where $\bPh_B^{(\ell)} = \C^{(\ell)} \odot \A^{(\ell)}$. Now let us consider the update of $\C$. The related optimization problem takes the form of
\begin{equation}
\min_{ \C \geq 0}  \| \tX^{(3)} - \bPh_C^{(\ell)} \C^T \|_F^2 + \mu   \| \C \|_F^2 + \nu  \| \C - \bar{\C}^{(\ell)} \|_F^2,
\label{eq:c_update}
\end{equation}
where $\bPh_C^{(\ell)}=\B^{(\ell)} \odot \A^{(\ell)}$, $\bar{\C}^{(\ell)}=( \P^{(\ell)} \circledast \Q^{(\ell)})\diag(\bbe^{(\ell)})$ and we define $\P^{(\ell)}(t,k) := S_k(t-1) $ and $\Q^{(\ell)}(t,k) := I_k(t-1) $.
Optimization problem~\eqref{eq:c_update} is also a nonnegative least squares problem. Therefore the updates for $\C$ are
\begin{subequations}
\begin{align}
& \widehat{\mathbf{C}}  = \argmin   \| \tX^{(3)} - \bPh_C^{(\ell)} \widehat{\mathbf{C}}  \|_2^2 + \mu   \| \widehat{ \mathbf{C}} \|_F^2  \\ 
&  \quad \quad \quad + \nu  \| \C - \bar{\C}^{(\ell)} \|_F^2+  \rho  \| \mathbf{C} - \widehat{\mathbf{C}}^T +  \mathbf{C}_d  \|_F^2 \notag \\
& \mathbf{C} =  \argmin_{\mathbf{C} \geq \mathbf{0} } \rho  \| \mathbf{C} - \widehat{\mathbf{C}}^T +  \mathbf{C}_d  \|_F^2 \\
& \mathbf{C}_d  =  \mathbf{C}_d + \mathbf{C} - \widehat{\mathbf{C}}^T
\end{align}
\label{eq:admmC}
\end{subequations}
We observed that running a few ADMM inner iterations ($\sim10$) for each factor suffices for the algorithm to produce satisfactory results.
\begin{algorithm}[t]
\caption{ \mname Method}
\label{alg:step}
\begin{algorithmic}
\STATE {\textbf{Input}: Tensor $\tX$, rank $K$, max. outer iterations $\rm{iters}_{\rm outer}$}, max. inner iterations $\rm{iters}_{\rm inner}$, gradient steps $\rm iters_{\rm grad}$, prediction window $L_o$
\REPEAT
\STATE Update $\A$ using~\eqref{eq:admmA} for $\rm{iters}_{\rm inner}$ iterations
\STATE Update $\B$ using~\eqref{eq:admmB} for $\rm{iters}_{\rm inner}$ iterations
\STATE Update $\C$ using~\eqref{eq:admmC} for $\rm{iters}_{\rm inner}$ iterations
\STATE Perform $\rm iters_{\rm grad}$ projected gradient steps for $\boldsymbol{\beta}$, $\boldsymbol{\gamma}$, $\mathbf{s}$, $\mathbf{i}$, using~\eqref{eq:derivebeta},~\eqref{eq:derivegamma},~\eqref{eq:derivs},~\eqref{eq:derivi}
\UNTIL{$\rm{iters}_{\rm outer}$ or validation RMSE increases}
\STATE Predict $L_o$ future slabs using Equation~\eqref{eq:prediction}
\end{algorithmic}
\end{algorithm}
By fixing the factor matrices, we have 
\begin{equation}
\begin{aligned}
\min_{\bbe, \bga, \mathbf{s}, \mathbf{i}} ~&  \sum_{k=1}^K \sum_{t=1}^L \left (c_{t,k} - \beta_k S_k(t-1) I_k(t-1) \right)^2    \\
 \text{s. t.} ~~& \bbe \geq \mathbf{0}, \bga \geq \mathbf{0}, \mathbf{s} \geq \mathbf{0},  \mathbf{i} \geq \mathbf{0}, \\
& S_k(t) = S_k(t-1) - \beta_k S_k(t-1) I_k(t-1), \\
& I_k(t) = I_k(t-1) + \beta_k S_k(t-1) I_k(t-1) \\
&\qquad \quad - \gamma_k I_k(t-1),  \\
& s_k = S_k(0), i_k = I_k(0) . 
\label{eq:subproblem}
\end{aligned}
\end{equation}
Both $S_k(t)$ and $I_k(t)$ are functions of $\bbe$ and $\bga$, and  are calculated in a recursive manner. Therefore the optimization problem w.r.t. $\bbe$ and $\bga$ is nonconvex. Optimization problem~\eqref{eq:subproblem} corresponds to $K$ independent one-dimensional curve fitting problems and since there are only $4$ optimization variables for each problem, we can use off-the-shelf curve fitting methods. Alternatively, we can perform a few projected gradient descent steps. Focusing on the $k$th subproblem we have
\begin{align}
f(\beta_k) = \nu \sum_{t=1}^L (c_{t,k} - \beta_k S_k(t-1) I_k(t-1))^2. 
\end{align}
The derivative of $f(\beta_k)$ w.r.t. $\beta_k$ is
\begin{equation}
\begin{aligned}
& \frac{\partial f}{\partial \beta_k} = -2\nu\sum_{t=1}^L(c_{t,k} - \beta_k S_k(t-1) I_k(t-1)) \times \\
& \Bigl(S_k(t-1)I_k(t-1) + \beta_k\frac{\partial S_k(t-1)}{\partial\beta_k}I_k(t-1) + \\
& \quad \quad \quad \quad \quad \quad \beta_k S_{t-1,k}\frac{\partial I_{t-1,k}}{\partial\beta_k} \Bigr).
\end{aligned}
\label{eq:derivebeta}
\end{equation}
Note that both $S_k(t)$ and $I_k(t)$ are recursive functions. Thus, their respective derivatives w.r.t. $\beta$ are computed recursively in $L$ steps. Similarly, for $\gamma_k$ we have 
\begin{equation}
\begin{aligned}
& \frac{\partial f}{\partial \gamma_k} =  - 2 \nu \sum_{t=1}^L (c_{t,k} - \beta_k S_k(t-1) I_k(t-1)) \times \\
&\left( \beta_k  \frac{\partial S_{t-1}}{\partial \gamma} I_{t-1} + \beta_k S_k(t-1) \frac{\partial I_k(t-1)}{\partial \gamma} \right).
\end{aligned}
\label{eq:derivegamma}
\end{equation}

Finally for $\mathbf{s}, \mathbf{i}$
\begin{equation}
\begin{aligned}
& \frac{\partial f}{\partial s_1} =  - 2 \nu \sum_{t=1}^L (c_{t,k} - \beta_k S_k(t-1) I_k(t-1)) \times \\
&\Bigl( \beta_k  \frac{\partial S_k(t-1)}{\partial s_1} I_k(t-1) + \beta_k S_k(t-1) \frac{\partial I_k(t-1)}{\partial s_1} \Bigr),
\end{aligned}
\label{eq:derivs}
\end{equation}
\begin{equation}
\begin{aligned}
& \frac{\partial f}{\partial i_1} = - 2 \nu \sum_{t=1}^L (c_{t,k} - \beta_k S_k(t-1) I_k(t-1)) \times \\
& \left( \beta_k  \frac{\partial S_k(t-1)}{\partial i_1} I_k(t-1) + \beta_k S_k(t-1) \frac{\partial I_k(t-1)}{\partial i_1} \right).
\end{aligned}
\label{eq:derivi}
\end{equation}
The overall procedure is summarized in Algorithm~\ref{alg:step}.

\begin{table*}[!t]
\centering
\resizebox{.9\textwidth}{!}{
\begin{tabular}{ c | c | c | c | c }
\toprule
& \multicolumn{2}{c|}{$L_o=10$} & \multicolumn{2}{c}{$L_o=15$}  \\
 Model & RMSE & MAE & RMSE & MAE  \\
 \midrule
  Mean & $304.1$ & $122.0$ &  $269.5$ & $108.5$\\
 SIR  &  $156.2$ &   $62.2$  &  $159.1$ &   $63.6$  \\
 SEIR &  $177.1$ &   $72.9$   &  $163.2$ &   $69.7$   \\
 LSTM (w/o feat.) &  $203.6$ &   $77.1$  &  $191.0$ &   $81.7$  \\
 LSTM (w/ feat.) &  $162.3$ &   $68.2$  &  $187.6$ &  $ 78.3$ \\
  STAN & $164.2$ & $61.1$ & $152.6$ & $ 61.8$  \\
 \mname ($\nu=0$) &  $149.2$ &   $61.5$  &  $152.8$ &   $ 66.9$  \\
 \mname &  $\mathbf{127.5}$ &   $\mathbf{55.6}$  &  $\mathbf{136.1}$ &   $\mathbf{61.7}$  \\
\bottomrule
\end{tabular}
\quad \quad
\begin{tabular}{c | c | c | c | c }
 \toprule
& \multicolumn{2}{c|}{$L_o=10$} & \multicolumn{2}{c}{$L_o=15$}  \\
 Model & RMSE & MAE & RMSE & MAE  \\
 \midrule
 Mean & $125.0$ & $77.0$ & $123.3$ & $77.1$\\
 SIR  &  $46.5$ &   $27.2$  &  $48.7$ &   $27.7$  \\
 SEIR &  $39.1$ &   $23.9$   &  $ 41.1$ &   $25.7$   \\
 LSTM (w/o feat.) & $45.6$ &   $23.6$  &  $54.8$ & $31.2$  \\
 LSTM (w/ feat.) &  $42.5$ &   $23.3$  &  $47.5$ & $26.8$ \\
  STAN & $30.6$ & $17.3$ & $42.8$ & $24.2$ \\
 \mname ($\nu=0$) &  $28.6$ &   $ 16.6 $  &  $46.8 $ &   $21.1 $  \\
 \mname &  $\mathbf{24.0}$ &   $\mathbf{15.1}$  &  $\mathbf{36.0}$ &   $\mathbf{18.0}$  \\
 \bottomrule
\end{tabular}
}
\caption{County-level prediction for new infections (left) and hospitalized patients (right). }
\label{table:county_level}
\end{table*}

\begin{table*}[!t]
\centering
\resizebox{.9\textwidth}{!}{
\begin{tabular}{ c | c | c | c | c }
 \toprule
& \multicolumn{2}{c|}{$L_o=10$} & \multicolumn{2}{c}{$L_o=15$}  \\
Model & RMSE & MAE & RMSE & MAE  \\
\midrule
 Mean & $309.0$ & $258.7$ &  $325.8$ & $273.1$\\
SIR  &  $186.1$ &   $133.8$  &  $186.9$ &   $134.5$  \\
SEIR &  $162.4$ &   $127.0$   &  $162.6$ &   $130.2$   \\
LSTM (w/o feat.) &  $187.5$ &   $138.1$  &  $419.7$ &   $356.0$  \\
LSTM (w/ feat.) &  $197.9$ &   $151.6$  &  $359.2$ &   $286.5$ \\
STAN & $\mathbf{74.1}$ & $\mathbf{60.1}$ & $\mathbf{100.5}$  & $79.6$ \\
\mname ($\nu=0$) &  $140.8 $ &   $ 104.0$  &  $ 127.8 $ &   $ 95.0 $  \\
\mname &  $117.8$ &   $89.8$  &  $107.3$ &   $\mathbf{79.4}$  \\
\bottomrule
\end{tabular}
\quad \quad
\begin{tabular}{ c | c | c | c | c }
 \toprule
& \multicolumn{2}{c|}{$L_o=10$} & \multicolumn{2}{c}{$L_o=15$}  \\
Model & RMSE & MAE & RMSE & MAE  \\
\midrule
Mean & $685.5$ & $553.9$ & $729.2$ & $586.0$\\
SIR  &  $343.4$ &   $252.4$  &  $367.8$ &   $266.0$  \\
SEIR &  $109.0$ &   $ 97.6$   &  $192.5$ &   $154.3$   \\
LSTM (w/o feat.) &  $280.9$ &   $187.5$  &  $416.1$ &   $308.7$  \\
LSTM (w/ feat.) &  $ 295.3$ &   $182.3$  &  $276.0$ &   $208.5$ \\
STAN & $100.3$ & $73.6$ & $177.7$ & $144.7$ \\
\mname ($\nu=0$) &  $ 118.3 $ &   $ 84.4 $  &  $ 126.6$ &   $\mathbf{75.2}$  \\
\mname &  $\mathbf{56.8}$ &   $\mathbf{43.9}$  &  $\mathbf{113.6}$ & $83.8$  \\
\bottomrule
\end{tabular}
}
\caption{State-level prediction for new infections (left) and hospitalized patients (right).}
\label{table:state_level}
\end{table*}

\begin{figure*}[t]
\centering
\begin{subfigure}{0.3 \textwidth}
\includegraphics[width=1\textwidth]{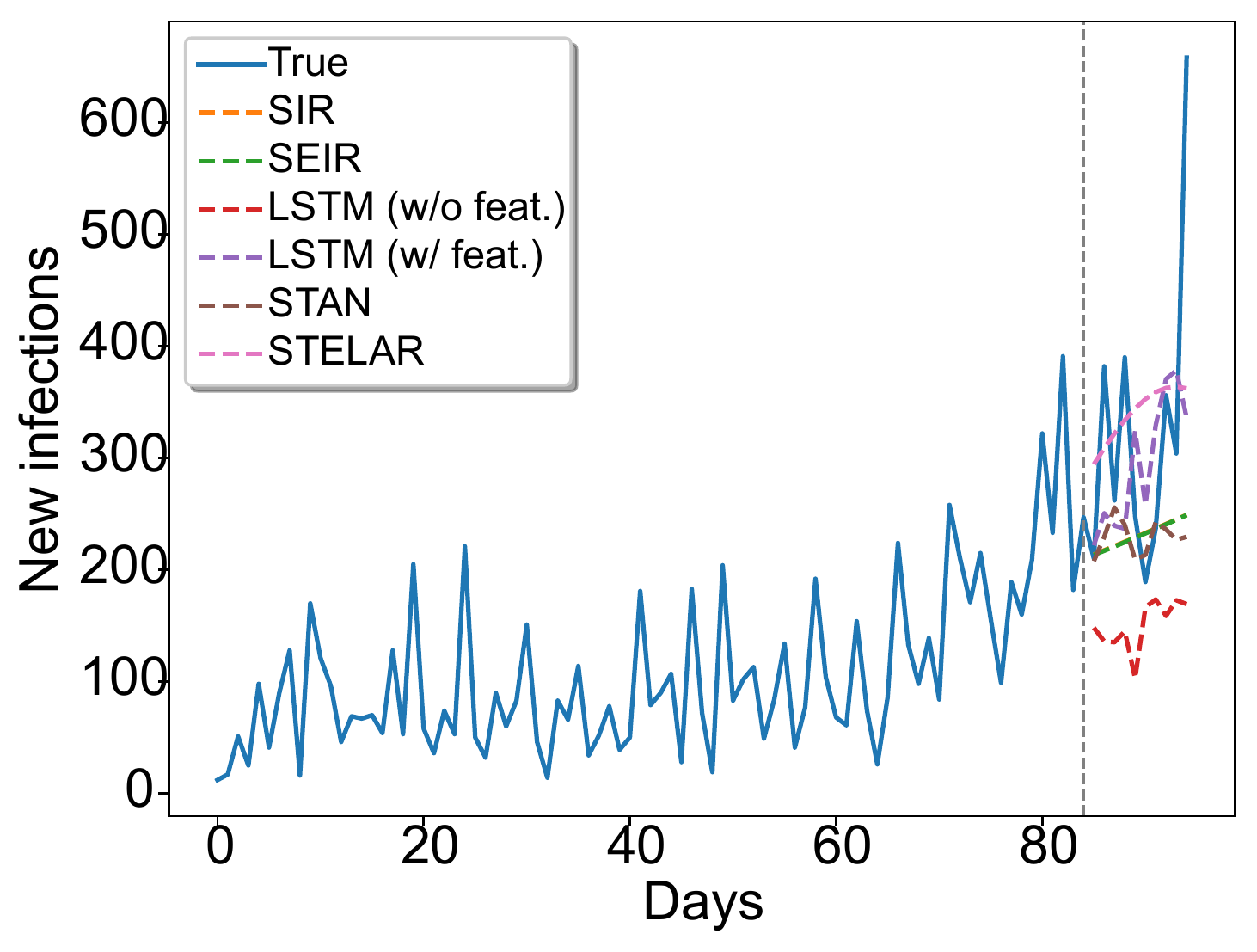}
\caption{Palm Beach, Florida}
\label{fig:example1}
\end{subfigure}
\begin{subfigure}{0.3 \textwidth}
\includegraphics[width=1\textwidth]{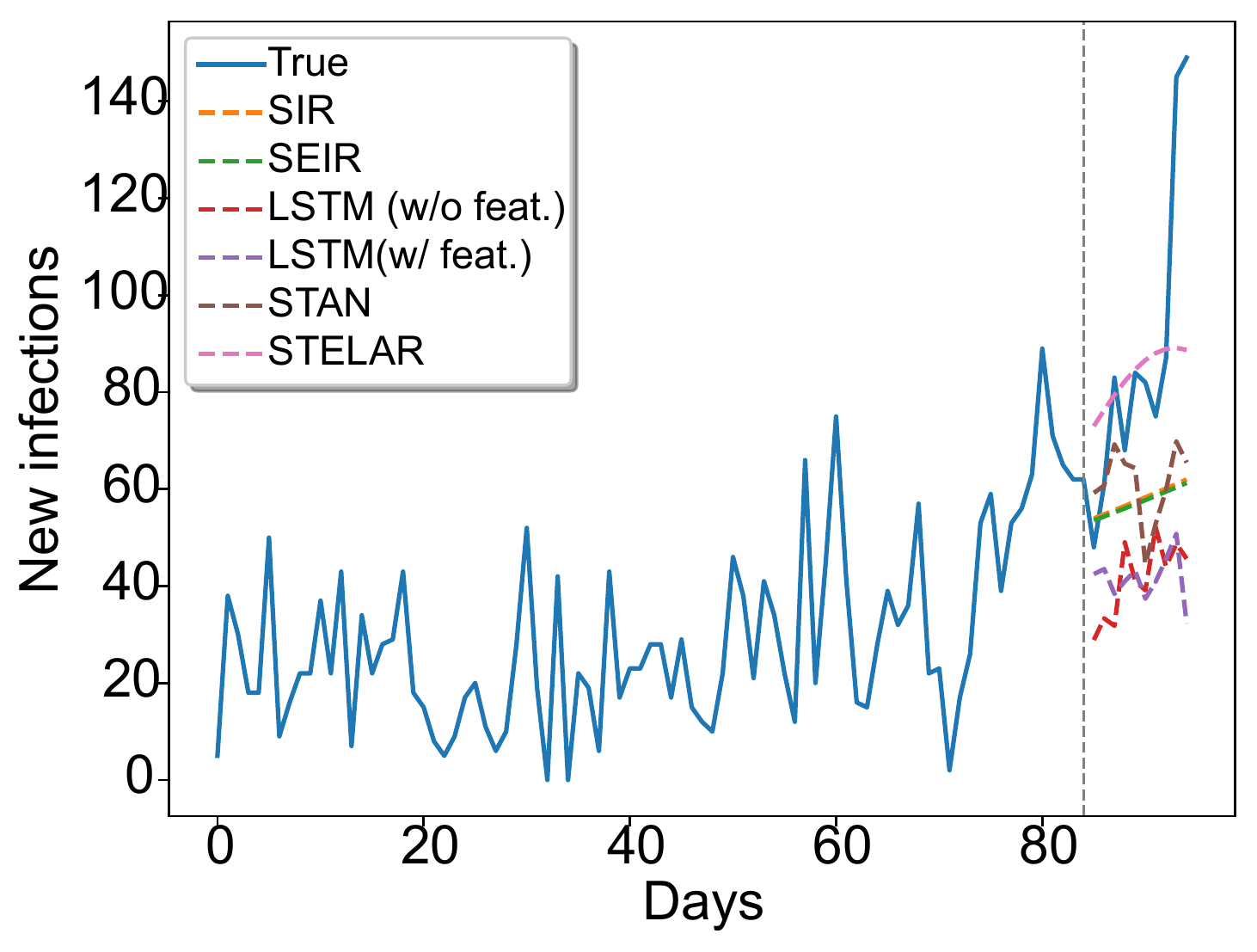}
\caption{Jefferson, Alabama}
\label{fig:example2}
\end{subfigure}
\begin{subfigure}{0.3 \textwidth}
\includegraphics[width=1\textwidth]{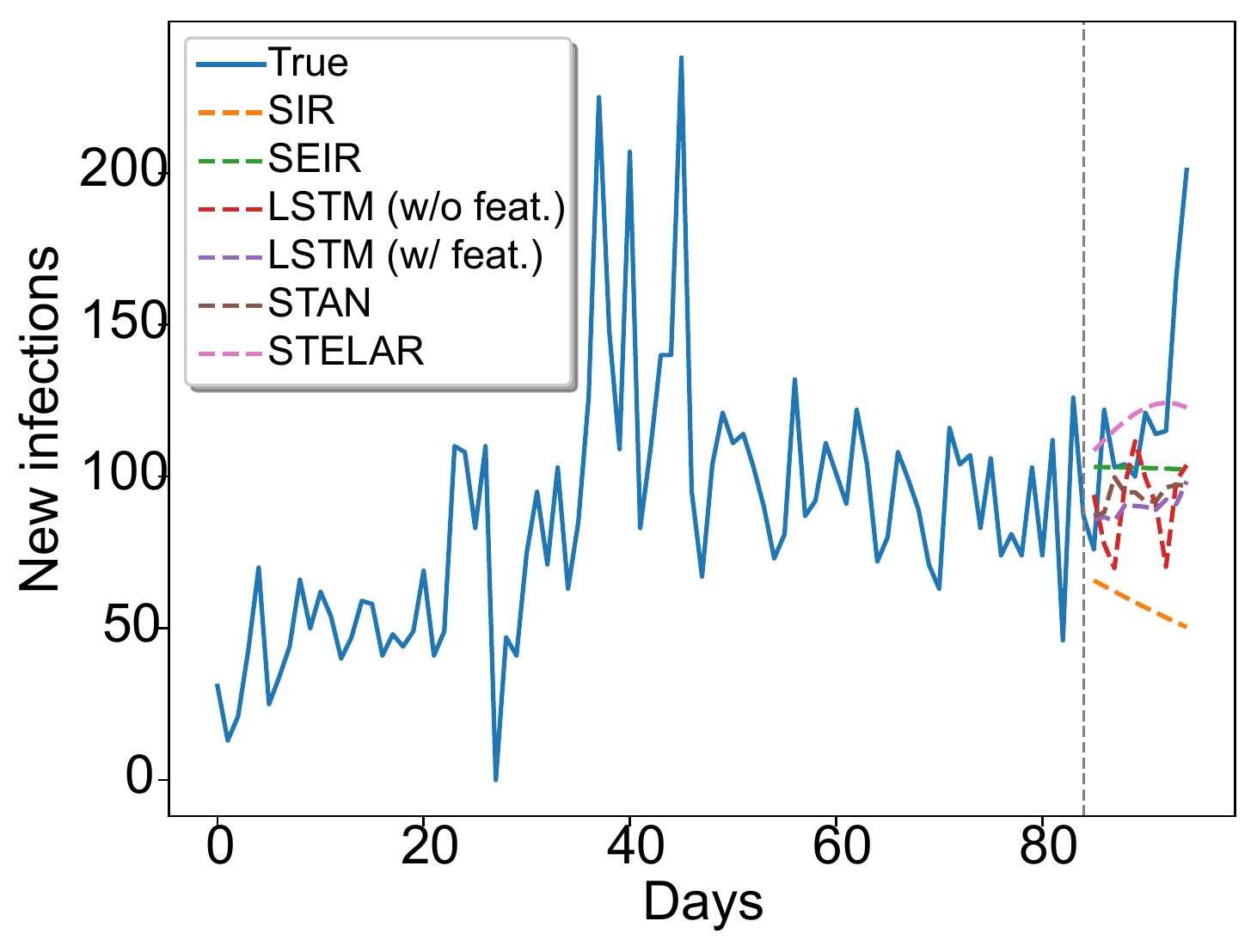}
\caption{Franklin, Ohio}
\label{fig:example3}
\end{subfigure}
\caption{$10$ days county-level prediction for new infections.}
\label{fig:county_examples}
\end{figure*}

\section{Experiments}
\label{experiment}

\subsection{Dataset and Baselines}
We use US county-level data from the Johns Hopkins University (JHU)~\cite{dong2020} and a large IQVIA patient claims dataset, which can be publicly accessible upon request. JHU data includes the number of active cases, confirmed cases and deaths for different counties in the US. The total number of counties was $133$. We use the reported active cases to compute the daily new infections. The claims dataset was created from 582,2748 claims from 732,269 COVID-19 patients from 03-24-2020 to 06-26-2020 (95 days). It contains the daily counts of $12$ International Classification of Diseases ICD-10 codes observed in each county and the Current Procedural Terminology (CPT) codes related to hospitalization and utilization of intensive care unit (ICU). The size of the constructed tensor is $133 \times 15 \times 95$. Using this data, we also construct a tensor which includes the aggregated data for $5$ states, New York, New Jersey, Massachusetts, Connecticut, Pennsylvania and is of size $5 \times 15 \times 95$. 

We perform two different experiments. Initially we use $85$ days for training and validation and use the remaining $L_o = 10$ as the test set. In the second experiment we use $80$ days for training and validation and $L_o = 15$ days for test. We compare our method against the following baselines:
\begin{enumerate}
\item \textbf{Mean}. We use the mean of the last $5$ days of the training set as our prediction.
\item \textbf{SIR}. The susceptible-infected-removed model.
\item \textbf{SEIR}. The susceptible-exposed-infected-removed epidemiological model.
\item \textbf{LSTM (w/o feat.)} LSTM model without additional features. We use one type of time series as our input.
\item \textbf{LSTM (w feat.)} LSTM with additional features. We use $15$ different time series as input.  
\item \textbf{STAN}~\cite{gao2020} A recently proposed GNN with attention mechanism.
\item \textbf{\mname} ($\nu=0$) The proposed method without regularization (two-step approach).
\end{enumerate}
We use the Root Mean Square Error (RMSE) and Mean Absolute Error (MAE) as our evaluation metrics.

\subsection{Results}

Table~\ref{table:county_level} shows the county-level results for $10$ and $15$ day prediction for daily new infections and hospitalized patients. For new infections and $L_o = 10$, our method achieves $18\%$ lower RMSE and $9\%$ lower MAE compared to the best performing baselines which are the SIR and STAN model respectively. When $L_o = 15$ our method achieves $10\%$ lower RMSE and the same MAE compared to the STAN model. For hospitalized patients and $L_o = 10$, our method achieves $21\%$ lower RMSE and $12\%$ lower MAE compared to the best performing baseline which is STAN. When $L_o = 15$ our method achieves $12\%$ lower RMSE and $25\%$ lower MAE compared to the best baseline. For state-level prediction of new infections, the best performing model is STAN but for hospitalized patients our method again outperforms all the baselines. Note that in all cases except one, joint optimization and SIR model fitting always improves the performance of our method compared to the two-step procedure.

Figure~\ref{fig:county_examples} shows some examples of county-level prediction. Figure~\ref{fig:example1} shows simple case where one can observe an increasing pattern of the new infections. Almost all models are able to capture this trend except for the LSTM (w/feat.) model. Figure~\ref{fig:example2} shows a more challenging scenario where it is not obvious if the curve will continue increasing but our method makes an accurate prediction, and is better than the baselines. Finally in~Figure~\ref{fig:example3} we observe an example where a peak was already observed in the past and therefore SIR and SEIR models fail. On the other hand, our method is again able to make reasonable predictions. 

\begin{table}[!t]
\centering
\resizebox{.4\textwidth}{!}{
\begin{tabular}{ c  | c | c  }
\toprule
Component 1 & Component 2 &  Component 3   \\
\midrule
New York      (NY) & L.A              (CA) & Nassau        (NY)          \\
Westchester   (NY) & Cook             (IL) & L.A           (CA)          \\
Nassau        (NY) & Milwaukee        (WI) & Essex         (NJ)          \\
Bergen        (NJ) & Fairfax          (VA) & Wayne         (MI)          \\
Miami-Dade    (FL) & Hennepin         (MN) & Oakland       (MI)          \\
Hudson        (NJ) & Montg.           (MD) & Middlesex     (NJ)          \\
Union         (NJ) & P. George's      (MD) & New York      (NY)          \\
Phila.        (PA) & Dallas           (TX) & Phila.        (PA)          \\
Passaic       (NJ) & Orange           (CA) & Cook          (IL)          \\
Essex         (NJ) & Harris           (TX) & Bergen        (NJ)          \\
\bottomrule
\end{tabular}
}
\caption{Counties that contribute more to each of the strongest $3$ rank-$1$ components of a rank-$30$ \mname model.
}
\label{table:counties}
\end{table}

\begin{figure}[!t]
\centering
\includegraphics[width=0.28\textwidth]{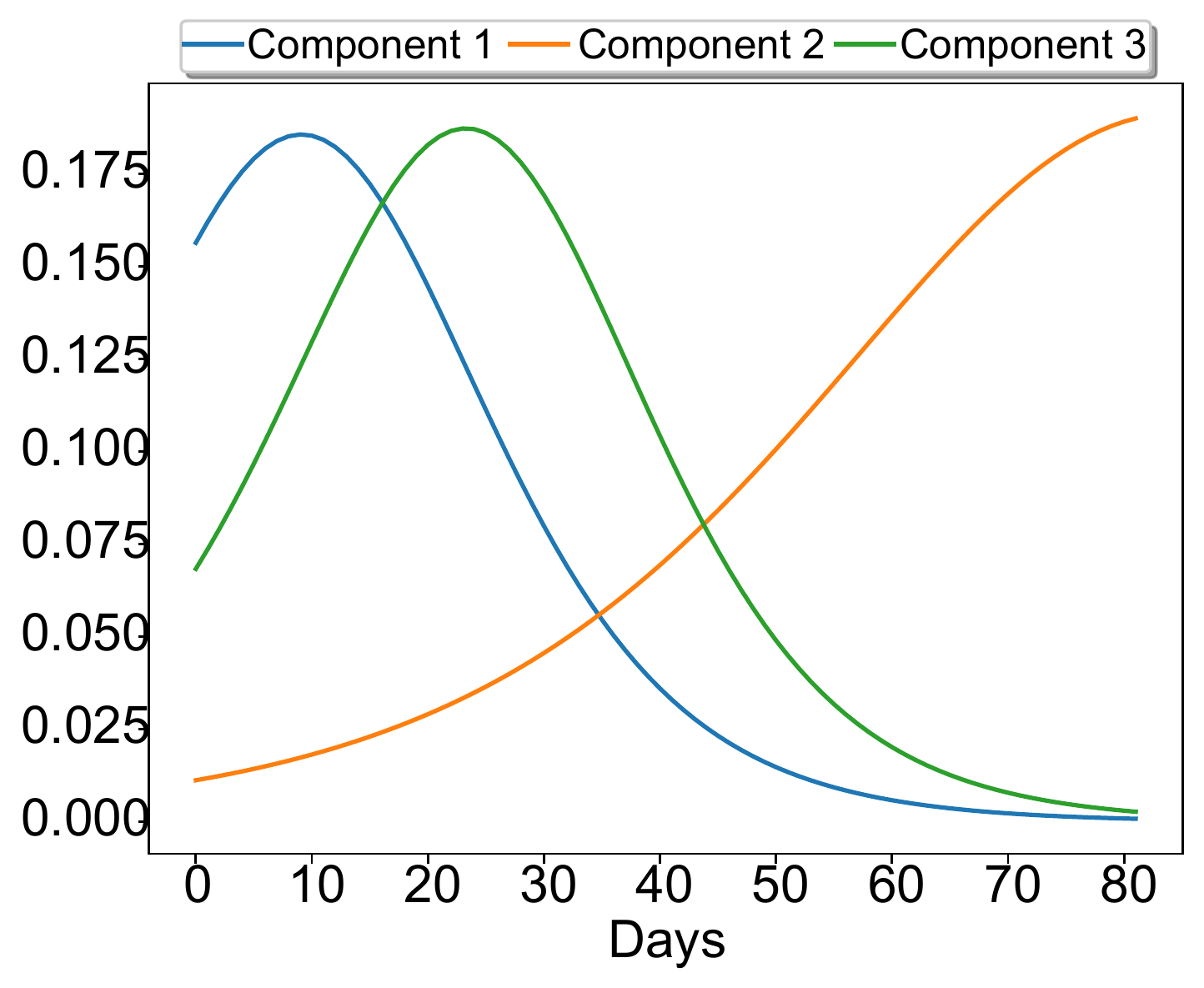}
\caption{The strongest $3$ temporal components of a rank-$30$ \mname model.}
\label{fig:interpretability}
\end{figure}

\begin{table}[!t]
\centering
\resizebox{.4\textwidth}{!}{
\begin{tabular}{ c  | c | c  }
\toprule
Component 1 & Component 2 &  Component 3    \\
\midrule
New infections         & New infections        & Hosp. patients  \\
 Hosp. patients        & Hosp. patients        & ICU patients    \\
 ICU patients          & ICU patients          & J96             \\
 J96                   & J96                   & N17             \\
 R09                   & R05                   & R06             \\
\bottomrule
\end{tabular}
}
\caption{Signals that contribute more to each of the strongest $3$ rank-$1$ components of a rank-$30$ \mname model. Definitions of ICD-10 codes: J96--Respiratory failure, not elsewhere classified; N17--Acute kidney failure; R05--Cough. R06--Abnormalities of breathing; R09--Other symptoms and signs involving the circulatory and respiratory system.}
\label{table:signals}
\end{table}
Finally, we demonstrate the ability of our model to produce interpretable results. We train our model on county-level data using $K=30$. After the algorithm converges we normalize each factor matrix such that each column has unit norm and absorb the scaling to a vector $w$. Using this vector, we extract the $3$ rank-$1$ components with the highest weights i.e., by selecting the corresponding columns of $\A, \B, \C$. Figure~\ref{fig:interpretability} depicts the $3$ temporal profiles (columns of factor $\C$) that corresponds to the selected rank-$1$ components. We observe that the different curves depict different stages of the disease transmission. For example, the $1$st component has captured an initial wave of the disease, the $2$nd has captured a more recent increasing trend and the $3$rd component is very similar to the first but shifted to the right. To gain some intuition regarding the results, we sort each column of factor matrices $\A$ (counties) and $\B$ (signals) and present the top $10$ counties and top $5$ signals that contribute to each one of the $3$ strongest components in Tables~\ref{table:counties},~\ref{table:signals}. We can clearly see that the counties which correspond to the early wave ($1$st component) are from the states of New York and New Jersey as we would expect. Also the $1$st component is mostly associated with new infections, hospitalized patients and ICU. On the other hand, the $3$rd component which is very similar to the $1$st is associated with hospitalized patients and ICU but not with new infections. Some counties appear in both the $1$st and $3$rd component which means that hospitalized patients and ICU cases started to increase (or getting reported) slightly after the new infections were reported. Finally, the second component depicts a later increase to the number of infections and hospitalized patients for some counties.
\section{Conclusion}
In this paper, we propose \mname a data efficient and interpretable method based on constrained nonnegative tensor factorization. Unlike  standard  tensor  factorization methods, our method enables long-term prediction of future slabs by incorporating latent epidemiological regularization. We demonstrated the ability of our method to make accurate predictions on real  county-  and state-level COVID-19 data. Our method achieves $18\%$ and $10\%$ lower RMSE compared to the best baseline, when predicting county-level daily new infections for $10$ and $15$ days-ahead respectively and  $21\%$ and $12\%$ lower RMSE when predicting county-level hospitalized patients.
\label{sec:imple}
\section*{Acknowledgments}
This work is in part supported by National Science Foundation award IIS-1704074, Army Research Office award ARO W911NF1910407, National Science Foundation award SCH-2014438, IIS-1418511, CCF-1533768, IIS-1838042, the National Institute of Health award NIH R01 1R01NS107291-01 and R56HL138415.

\bibliography{main}

\begin{thebibliography}{21}
\providecommand{\natexlab}[1]{#1}
\providecommand{\url}[1]{\texttt{#1}}
\providecommand{\urlprefix}{URL }
\expandafter\ifx\csname urlstyle\endcsname\relax
  \providecommand{\doi}[1]{doi:\discretionary{}{}{}#1}\else
  \providecommand{\doi}{doi:\discretionary{}{}{}\begingroup
  \urlstyle{rm}\Url}\fi

\bibitem[{Acar, Dunlavy, and Kolda(2009)}]{acar2009}
Acar, E.; Dunlavy, D.~M.; and Kolda, T.~G. 2009.
\newblock Link prediction on evolving data using matrix and tensor
  factorizations.
\newblock In \emph{2009 IEEE International Conference on Data Mining Workshops
  (ICDMW)}, 262--269.

\bibitem[{Araujo, Ribeiro, and Faloutsos(2017)}]{de2017}
Araujo, M.~R.; Ribeiro, P. M.~P.; and Faloutsos, C. 2017.
\newblock {TensorCast}: Forecasting with context using coupled tensors.
\newblock In \emph{2017 IEEE International Conference on Data Mining (ICDM)},
  71--80.

\bibitem[{Chimmula and Zhang(2020)}]{chimmula2020}
Chimmula, V. K.~R.; and Zhang, L. 2020.
\newblock Time series forecasting of {COVID}-19 transmission in Canada using
  {LSTM} networks.
\newblock \emph{Chaos, Solitons \& Fractals} 135.

\bibitem[{Cooke and Van Den~Driessche(1996)}]{cooke1996}
Cooke, K.~L.; and Van Den~Driessche, P. 1996.
\newblock Analysis of an {SEIRS} epidemic model with two delays.
\newblock \emph{Journal of Mathematical Biology} 35(2): 240--260.

\bibitem[{Dong, Du, and Gardner(2020)}]{dong2020}
Dong, E.; Du, H.; and Gardner, L. 2020.
\newblock An interactive web-based dashboard to track {COVID}-19 in real time.
\newblock \emph{The Lancet infectious diseases} 20(5): 533--534.

\bibitem[{Dunlavy, Kolda, and Acar(2011)}]{dunlavy2011}
Dunlavy, D.~M.; Kolda, T.~G.; and Acar, E. 2011.
\newblock Temporal link prediction using matrix and tensor factorizations.
\newblock \emph{ACM Transactions on Knowledge Discovery from Data (TKDD)} 5(2):
  1--27.

\bibitem[{Gabay and Mercier(1976)}]{gabay1976}
Gabay, D.; and Mercier, B. 1976.
\newblock A dual algorithm for the solution of nonlinear variational problems
  via finite element approximation.
\newblock \emph{Computers \& Mathematics with Applications} 2(1): 17--40.

\bibitem[{Gao et~al.(2020)Gao, Sharma, Qian, Glass, Spaeder, Romberg, Sun, and
  Xiao}]{gao2020}
Gao, J.; Sharma, R.; Qian, C.; Glass, L.~M.; Spaeder, J.; Romberg, J.; Sun, J.;
  and Xiao, C. 2020.
\newblock STAN: Spatio-Temporal Attention Network for Pandemic Prediction Using
  Real World Evidence.
\newblock \emph{arXiv preprint arXiv:2008.04215} .

\bibitem[{Harshman(1970)}]{Harshman1970}
Harshman, R.~A. 1970.
\newblock Foundations of the {PARAFAC} procedure: Models and conditions for an
  {"explanatory"} multimodal factor analysis.
\newblock \emph{UCLA Working Papers Phonetics} 16: 1--84.

\bibitem[{He, Peng, and Sun(2020)}]{he2020}
He, S.; Peng, Y.; and Sun, K. 2020.
\newblock {SEIR} modeling of the {COVID}-19 and its dynamics.
\newblock \emph{Nonlinear Dynamics} 1--14.

\bibitem[{Hochreiter and Schmidhuber(1997)}]{hochreiter1997}
Hochreiter, S.; and Schmidhuber, J. 1997.
\newblock Long short-term memory.
\newblock \emph{Neural computation} 9(8): 1735--1780.

\bibitem[{Huang, Sidiropoulos, and Liavas(2016)}]{huang2016}
Huang, K.; Sidiropoulos, N.~D.; and Liavas, A.~P. 2016.
\newblock A flexible and efficient algorithmic framework for constrained matrix
  and tensor factorization.
\newblock \emph{IEEE Transactions on Signal Processing} 64(19): 5052--5065.

\bibitem[{Kapoor et~al.(2020)Kapoor, Ben, Liu, Perozzi, Barnes, Blais, and
  O'Banion}]{kapoor2020}
Kapoor, A.; Ben, X.; Liu, L.; Perozzi, B.; Barnes, M.; Blais, M.; and O'Banion,
  S. 2020.
\newblock Examining {COVID}-19 Forecasting using Spatio-Temporal Graph Neural
  Networks.
\newblock \emph{arXiv preprint arXiv:2007.03113} .

\bibitem[{Kermack and McKendrick(1927)}]{kermack1927}
Kermack, W.~O.; and McKendrick, A.~G. 1927.
\newblock A contribution to the mathematical theory of epidemics.
\newblock \emph{Proceedings of the royal society of london. Series A,
  Containing papers of a mathematical and physical character} 115(772):
  700--721.

\bibitem[{Kipf and Welling(2017)}]{kipf2017}
Kipf, T.~N.; and Welling, M. 2017.
\newblock Semi-Supervised Classification with Graph Convolutional Networks.
\newblock In \emph{2017 International Conference on Learning Representations
  (ICLR)}.

\bibitem[{Papalexakis, Faloutsos, and Sidiropoulos(2016)}]{papalexakis2016}
Papalexakis, E.~E.; Faloutsos, C.; and Sidiropoulos, N.~D. 2016.
\newblock Tensors for data mining and data fusion: Models, applications, and
  scalable algorithms.
\newblock \emph{ACM Transactions on Intelligent Systems and Technology (TIST)}
  8(2): 1--44.

\bibitem[{Sidiropoulos et~al.(2017)Sidiropoulos, De~Lathauwer, Fu, Huang,
  Papalexakis, and Faloutsos}]{Sidiropoulos2017}
Sidiropoulos, N.~D.; De~Lathauwer, L.; Fu, X.; Huang, K.; Papalexakis, E.~E.;
  and Faloutsos, C. 2017.
\newblock Tensor Decomposition for Signal Processing and Machine Learning.
\newblock \emph{IEEE Transactions on Signal Processing} 65(13): 3551--3582.

\bibitem[{Toda(2020)}]{toda2020}
Toda, A.~A. 2020.
\newblock Susceptible-infected-recovered ({SIR}) dynamics of {COVID}-19 and
  economic impact.
\newblock \emph{arXiv preprint arXiv:2003.11221} .

\bibitem[{Tomar and Gupta(2020)}]{tomar2020}
Tomar, A.; and Gupta, N. 2020.
\newblock Prediction for the spread of {COVID}-19 in India and effectiveness of
  preventive measures.
\newblock \emph{Science of The Total Environment} 18.

\bibitem[{Wang, Chen, and Marathe(2019)}]{wang2019}
Wang, L.; Chen, J.; and Marathe, M. 2019.
\newblock DEFSI: Deep learning based epidemic forecasting with synthetic
  information.
\newblock In \emph{Proceedings of the AAAI Conference on Artificial
  Intelligence}, volume~33, 9607--9612.

\bibitem[{Yang et~al.(2020)Yang, Zeng, Wang, Wong, Liang, Zanin, Liu, Cao, Gao,
  Mai et~al.}]{yang2020}
Yang, Z.; Zeng, Z.; Wang, K.; Wong, S.-S.; Liang, W.; Zanin, M.; Liu, P.; Cao,
  X.; Gao, Z.; Mai, Z.; et~al. 2020.
\newblock Modified SEIR and AI prediction of the epidemics trend of {COVID}-19
  in China under public health interventions.
\newblock \emph{Journal of Thoracic Disease} 12(3): 165.

\end{thebibliography}

\end{document}